\documentclass{article}

\usepackage{arxiv}

\usepackage[utf8]{inputenc} 
\usepackage[T1]{fontenc}    
\usepackage{hyperref}       
\usepackage{url}            
\usepackage{booktabs}       
\usepackage{amsfonts}       
\usepackage{nicefrac}       
\usepackage{microtype}      
\usepackage{lipsum}
\usepackage{graphicx}
\graphicspath{ {./images/} }
\usepackage{listings}
\usepackage{wrapfig}
\usepackage{subcaption}
\usepackage{array}

\title{Scientific Explanations in Health Sciences: \\ Causality, Trust, and Epistemic Adequacy}

\author{
 Martina Mattioli \\
  College of Mathematical Medicine\\
  Zhejiang Normal University\\
  Jinhua, China \\
  \texttt{martinamattioli2022@gmail.com} \\
   \And
 Marcello Pelillo \\
  DAIS, Ca' Foscari University of Venice, Italy\\
  College of Mathematical Medicine, Zhejiang Normal University, China\\
  European Centre for Living Technology, Italy\\
  \texttt{pelillo@unive.it} \\
}

\begin{document}
\maketitle
\begin{abstract}
Medical Artificial Intelligence (AI) is widely expected to transform clinical practice, yet the decision-making processes of many Machine Learning (ML) models remain opaque. Explainability has been advanced as a partial remedy to clarify why AI generates predictions, particularly in high-stakes contexts. Despite ongoing efforts, debates on what constitutes an adequate medical explanation remain unsettled.
Yet, explanation has long been a central topic of inquiry in the philosophy of science and medicine. The insights developed in these fields, however, have been largely overlooked in contemporary explainable AI (XAI) research, leaving its foundational assumptions insufficiently examined. 
To address this gap, this paper develops a critical review at the intersection of philosophy of science and XAI. It examines prevailing accounts of what counts as an explanation in the health sciences and assesses their adequacy for informing XAI in medicine, arguing that they provide necessary conditions for a philosophically grounded approach to explainability in this domain.
Building on this foundational philosophical literature, the discussion identifies three central axes of analysis: the role of causality in medical reasoning, the epistemic and relational dimensions of medical trust, and the criteria of explanatory adequacy as shaped by the pragmatic needs of diverse stakeholders. 
By integrating philosophical analysis with current developments in medical AI, the paper outlines principles for designing XAI systems that offer explanations that are not only epistemically robust but also aligned with the epistemic and practical requirements of clinical decision-making, shaping ongoing debates in medical XAI toward underexplored conceptual foundations.
\end{abstract}


\section{Introduction}
The scope of AI applications is rapidly expanding. In particular, medical AI is expected to profoundly transform the practice of medicine~\cite{rajpurkar2022ai}. 
Yet, the black box nature of many ML algorithms poses persistent challenges, as their decision-making processes are often opaque and poorly understood~\cite{tjoa2020survey}. 
This opacity is especially problematic in medicine, where algorithmic decisions can directly impact patient outcomes, often determining matters of life and death~\cite{tjoa2020survey,nyrup2022explanatory}.
To address this lack of transparency and the sensitive nature of medical decisions, explainability has been introduced as a partial remedy to clarify how and why artificial systems generate predictions~\cite{ribeiro2016should}.
However, it is widely acknowledged that the field of XAI lacks conceptual clarity, as multiple definitions and frameworks coexist in the absence of consensus on what constitutes an adequate explanation~\cite{Paez19pragmatic,Mattioli2024UnderstandingXAI}. These difficulties become especially salient in the medical domain, where disagreement persists over a shared philosophical account of explanation, particularly when XAI is integrated into diagnostic systems and clinical practice~\cite{boge2025causality,london2019artificial,duran2021dissecting}. 
For instance, the literature diverges on whether medical explanations should meet scientific and causal standards or whether performance-oriented criteria may suffice~\cite{boge2025causality,london2019artificial}.
As a result, the debate remains open, with no explicit agreement on the foundations of explanation in the context of medical XAI. 

Conversely, the field of philosophy of science encompasses refined accounts of explanation, many of which are connected to various views of causality, trust, explanatory adequacy, and other topics of interest in XAI, thereby shedding new light on these debates~\cite{Mattioli2024UnderstandingXAI}. 
Besides these overarching theories, philosophers have specifically examined the nature of explanations in medicine, investigating how they interact with clinical reasoning, medical trust, and other pragmatic concerns~\cite{russo2007interpreting,clark2002trust,schaffner1993discovery}. 
These studies not only address the conditions under which explanatory claims constitute a valid explanation, but also the epistemic and normative requirements that explanations must meet in clinical contexts.
The philosophical literature thus provides a comprehensive conceptual framework for understanding XAI in medicine. It offers analytical perspectives for explanations that are scientifically informed and clinically meaningful, while also supporting both decision-making and trust in AI systems. Clarifying what medicine requires of its explanations and engaging with the philosophical dimension of explanation are, we argue, essential conditions for identifying XAI foundations and assessing whether any XAI approach can meet clinical and knowledge requirements. Indeed, medical explanations are not simply technical tools but epistemic and relational practices that, in many clinical contexts, aim at intervention~\cite{boge2025causality}.

Building on these concerns, this paper makes three distinctive contributions: first, it provides a critical review of the main concepts and debates surrounding medical explanations in AI, determining the conceptual and practical gaps between the expectations established by philosophical accounts of explanation and the current capabilities of AI explanations in healthcare.
Secondly, the paper identifies and analyzes three core areas, drawn from the philosophical literature on medical explanation, to support XAI. These address conceptions of~\textit{causality}, the challenges in establishing~\textit{medical trust}, and the~\textit{adequacy of explanations} in relation to~\textit{pragmatic factors} and different stakeholders. 
Finally, we advance epistemically grounded design principles that show how philosophical perspectives can inform medical explanations that are not only epistemically and philosophically robust but also practically relevant to the requirements of medical contexts. 
To achieve this, in Section~\ref{sec:landscape-xai}, we examine XAI in medicine, evaluating current practices, principles, and ongoing debates. 
Section~\ref{sec:phil-med-ex} focuses on the philosophical perspectives on medical explanation, with particular attention to the tension between causality and correlation, the role of trust, and criteria for~\textit{bona fide} explanations. 
Finally, Section~\ref{sec:bridging-medical} connects philosophy and the XAI field, identifying design principles for medical XAI to prioritize causal reasoning, stakeholder-specific needs, and the assessment of trust alongside accuracy.

\section{The Current Debates in Medical XAI}
\label{sec:landscape-xai}
The existing XAI framework in healthcare encompasses a range of methods to reduce the opacity of black box models and make AI decisions intelligible to clinicians, patients, and regulators~\cite{rajpurkar2022ai}. While methodological advances have multiplied, the field remains defined by ongoing discussions concerning the principles that should govern explainability and the trade-offs it entails. 

A central debate concerns the~\textit{balance between accuracy and interpretability}. 
Alex J. London~\cite{london2019artificial} has argued that demands for explainability in medicine may be misplaced, since medical practice often proceeds successfully in the absence of robust causal models, relying instead on empirically validated associations. For him, accuracy should be prioritized over providing rationales. Requiring explanations may, in his view, risk undermining performance and delaying the adoption of clinically beneficial systems.
In contrast, Christian Herzog~\cite{herzog2022ethical}, and later Florian J. Boge and Axel Mosig~\cite{boge2025causality}, contend that London’s account underestimates the epistemic and moral importance of explainability. They argue that trust cannot be grounded in accuracy alone but must also involve justification, transparency, and accountability within socio-technical systems~\cite{herzog2022ethical}. Moreover, medical XAI should meet scientific standards of explanation, that is, explanations that are testable, falsifiable, and tied to~\textit{causal hypotheses}~\cite{boge2025causality}. 
The tension between these positions has generated further debate on whether interpretability necessarily comes at the expense of accuracy. While London~\cite{london2019artificial} emphasizes such a trade-off, recent works demonstrate that inherent interpretable or causal models can achieve state-of-the-art performance and reveal hidden flaws in opaque black box systems~\cite{herzog2022ethical,rudin2019stop}. Consequently, the issue shifts from prioritizing accuracy or explainability to identifying design strategies that enable them both. 

This discussion aligns with the broader principle of~\textit{causability}~\cite{holzinger2019causability}, which is advanced in the medical XAI literature as the degree to which an explanation supports causal understanding for a human expert. This concept is proposed as a~\textit{desideratum}, distinct from causality or simple technical explainability, and is assessed in human-centered interactions. Causability concerns the quality of interaction between humans and AI, specifically whether an explanation enables clinicians to reason causally, align model outputs with domain knowledge, and pose counterfactual questions. This approach connects explanatory methods to practical decision-making by emphasizing that effective explanations enhance clinicians’ ability to evaluate potential outcomes of different interventions or treatment options~\cite{holzinger2019causability}.
Additional epistemic and ethical concerns include the~\textit{automation bias} and the risk of following and trusting AI recommendations when these systems demonstrate high performance and accuracy~\cite{herzog2022ethical,bjerring2021artificial}. This dynamic may undermine clinician oversight and patient-centered deliberation unless explanatory interfaces and validation protocols are rigorously implemented~\cite{herzog2022ethical}. 
These debates demonstrate that XAI in healthcare is not only a technical endeavor but also a philosophical one, positioned at the intersection of epistemology, ethics, and clinical intervention~\cite{boge2025causality,herzog2022ethical,holzinger2019causability}. 
Nonetheless, these discussions reveal a conceptual asymmetry: while the medical XAI literature invokes concepts such as trust, causality, and explanatory adequacy, it is often narrowly oriented on reductionist interpretations of these notions and rarely engages with the philosophical~\textit{substrata} in which they have been most rigorously defined. This motivates us to examine how these principles intersect with, and can be complemented by, epistemological discussions of medical explanation.

\section{Philosophical Views on Medical Explanation}
\label{sec:phil-med-ex}
We identified three thematic areas that shape the philosophical debates on medical explanation and that are relevant to XAI in healthcare: causality, medical trust, and~\textit{bona fide} explanations.

\subsection{Causality in Medical Reasoning}
Causal reasoning is widely recognized as fundamental to medicine. However, its exact nature is deeply contested and subject to multiple interpretations~\cite{russo2007interpreting,Salmon1990}. 
From one perspective, medicine draws on the natural sciences to identify general causal laws that account for disease occurrence and progression. On this view, medical explanations involve comprehension of intervening relationships, as effective therapy depends on pinpointing where interventions alter outcomes~\cite{boge2025causality}. 
Nevertheless, others argue that no single model of causality is sufficient for medical reasoning~\cite{russo2007interpreting}. 
Federica Russo and Jon Williamson~\cite{russo2007interpreting} claim that probabilistic evidence alone cannot justify causal claims, as probability distributions may reflect correlations without underlying mechanisms. Conversely, mechanistic explanations are insufficient without empirical validation. Their theory of causality, therefore, requires a ``duality of evidence'': causal claims are warranted only when both mechanistic pathways and probabilistic dependencies are established.
Furthermore, Ren Zong Qiu~\cite{qiu1989models} asserts that most diseases are not the outcome of a single chain of causes but of multiple interacting factors, such as genetic, environmental, social, and behavioral, that form a ``web of causation.'' 
Unlike mono-causal models that seek a single cause, this framework maintains that understanding the etiology of complex diseases, such as cancer or cardiovascular conditions, requires mapping the interconnections among numerous contributing factors. The value of this approach lies in its capability to inform preventive strategies: if causation is distributed across a web, interventions can be targeted at multiple nodes rather than seeking a unique cause.
This methodological shift reflects aspects of Chinese philosophy, whose conceptions of causality diverge significantly from Western ``billiard-ball'' mechanistic models. In this scheme, explanation is achieved by situating phenomena within ordered patterns of interdependence, rather than by subsuming them under general causal laws~\cite{defoort2017causation}. Indeed, Chinese medicine does not isolate single causal agents; instead, it situates illness within dynamic interactions among bodily systems and environmental factors~\cite{defoort2017causation,cheng1976model}.

Kenneth Schaffner~\cite{schaffner1983explanation} further refines the discussions on causality by emphasizing the probabilistic nature of biomedical explanation. He argues that while deterministic laws may apply to specific physical processes, medical phenomena often exhibit stochastic variability, making strict necessity an unrealistic standard~\cite{schaffner1983explanation}. According to him, statistical causation, while less robust than deterministic causation, retains explanatory value. 
Importantly, he distinguishes between mere statistical correlations and genuine probabilistic causal claims: the former describe regularities, while the latter articulate structured tendencies that can guide both scientific understanding and clinical decision-making.
From a broader philosophical perspective, this flexibility illustrates why medicine resists being reduced to a unique conception of causality. Mechanistic, probabilistic, and relational frameworks are applied according to whether the clinical priority is explanation, prediction, or intervention. In this sense, medical reasoning embodies pragmatic pluralism. What counts as a valid explanation depends on the clinical task, the level of analysis, and the type of decision at stake. 
Indeed, this pragmatic orientation is reflected in clinical practice. 
In emergency care, for instance, physicians focus on identifying proximal and modifiable causes that can be addressed to improve patient outcomes. In contrast, preventive medicine emphasizes distal etiological factors, including lifestyle and genetic predispositions, which influence disease risk at the population level~\cite{rizzi1994causal}. This pragmatic perspective underscores the dual requirements of medical causality: interventions must be actionable for individual patients and generalizable to broader contexts.
Philosophically, this pluralism challenges reductionist accounts of causality, demonstrating that medicine employs different causal models in response to the epistemic and practical needs of the situation. These accounts suggest that XAI methods in medicine cannot assume a single causal model. Instead, they need to present causal information in multiple forms. A feature attribution method may suffice where probabilistic risk communication is the goal; a counterfactual or mechanistic explanation is required where an intervention must be justified. 
Thus, philosophical accounts of medical causality provide multiple context-dependent criteria for evaluating XAI solutions.

\subsection{Trust and Clinical Adoption}
Trust is often identified as essential for the clinical adoption of medical technologies, including AI-based decision-support systems~\cite{rawal2025causality,nicolson2025human}. Patients are required to accept physicians' guidance, while physicians are expected to rely on the outputs of technological systems whose internal functioning may not always be transparent. 
As a result, trust is fundamentally shaped by the inherent asymmetry between patients, physicians~\cite{clark2002trust}, and, lately, AI. Medical interactions involve urgency, patient vulnerability, and the necessity to make significant decisions under uncertainty. This combination generates a form of trust that is~\textit{sui generis}, creating the basis of the fiduciary relationship between doctor and patient~\cite{clark2002trust}. 
This relationship is characterized by a justified expectation of the physician's skills and commitment, accepted under conditions of~\textit{inherent uncertainty} and risk. 
Stephen Holland and David Stocks~\cite{holland2017trust} propose a tripartite framework for understanding trust in medical contexts, distinguishing between reliance, specific trust, and general trust. Reliance refers to everyday expectations of competence and reliability: patients rely on physicians to apply professional standards just as they rely on institutional procedures. Specific trust emerges in situations of heightened vulnerability, requiring patients to believe that physicians will consider their individual needs, particularly when treatment decisions are uncertain. General trust entails an unconditional sense of safety and is typically reserved for deeply vulnerable situations.
This framework aligns with an analysis of trust in which ``reliance'' stems from perceived certainty from past performance, whereas ``trust'' involves a conscious acceptance of risk and is fundamentally agent-related, concerning the physician's behavior rather than a particular result~\cite{wolfensberger2019trust}.

The introduction of AI systems in clinical settings introduces an additional layer into this structure. Physicians must evaluate whether to trust the outputs generated by AI models, while patients must, in turn, decide whether to trust their physicians’ reliance on such systems. 
Opaque black box models that cannot provide intelligible reasoning risk to erode this chain of trust, since they contradict the principles of evidence-based medicine and the professional responsibility of physicians~\cite{kundu2021ai}. This erosion arises from the fact that trust, to be accountable, needs to be justified~\cite{wolfensberger2019trust}. For a clinician, establishing trust in an AI system necessitates sufficient reason to believe in its epistemic trustworthiness, which opacity directly undermines.
Thus, explanation plays a pivotal role in sustaining trust by conveying information and structuring the relational dynamics between human understanding and machines~\cite{ribeiro2016should,Paez19pragmatic,rawal2025causality}. Explanations should be understood as a social practice, co-constructed between explainer and explainee~\cite{rohlfing2020explanation}. 
Consequently, trust in medical AI will depend not only on the internal accuracy of models or the fidelity of explanations but also on how explanations are tailored to the needs of doctors and patients. Plausible yet unfaithful explanations may produce misplaced confidence~\cite{agarwal2024faithfulness}, which is particularly dangerous in high-stakes medical contexts. This demand, therefore, is not merely satisfied by interpretability or high accuracy but necessitates faithful and context-sensitive explanations that account for the specific vulnerabilities and fiduciary obligations of patients and clinicians. XAI evaluation frameworks that treat trust as equivalent to user satisfaction or perceived plausibility, therefore, miss what is philosophically and clinically at stake. For clinical adoption, we argue, trust must be cultivated at three levels: patients need to trust their physicians to interpret AI outputs responsibly, physicians must trust AI tools to provide reliable and transparent reasoning, and institutions must trust that these systems can be integrated without compromising ethical and professional standards. 

\subsection{What Counts as a~\textit{Bona Fide} Medical Explanation?}
The criteria for what makes a medical explanation ``good'' are central not only to philosophy, but also to the development of XAI~\cite{Salmon1990,buijsman2022defining}. 
In contrast to the natural sciences, where explanatory adequacy is often tied to generality, causality, or lawlikeness~\cite{Salmon1990}, medicine serves heterogeneous objectives: understanding disease mechanisms, predicting outcomes, guiding interventions, and providing intelligibility to patients~\cite{boge2025causality}. 
Paul Thagard~\cite{thagard1998explaining} contends that adequate explanations in medicine should be understood as explanatory networks that combine multiple strands of evidence, statistical, mechanistic, and experiential, into a coherent whole. This network-based perspective is well-suited to the complexity of medical conditions, which rarely have a single identifiable cause but rather emerge from the interaction of biological and environmental factors. Similarly, Qiu~\cite{qiu1989models} and Russo and Williamson~\cite{russo2007interpreting} emphasize that diseases cannot be reduced to single accounts of causality but must be understood through webs of interacting conditions. These perspectives underscore the complexity of explanatory adequacy in medicine: a ``good'' explanation requires capturing systemic interdependencies.

At the same time, explanatory adequacy is not only a matter of epistemic content but also of pragmatic function, shaped by the needs of the explainee~\cite{nyrup2022explanatory}. 
For physicians, adequate explanations provide causal detail to justify an intervention. For patients, they offer narratives that relate symptoms to understandable causes using everyday language. 
Staffan Norell~\cite{norell1984models} further argues that medical explanations must be simple, plausible, and fit the complexity of empirical observations. Simplicity is essential because overly elaborate explanations can obscure rather than clarify. Plausibility ensures that explanations remain anchored in empirical regularities. Addressing data complexity prevents the adoption of oversimplified mono-causal accounts. 
Explanatory adequacy requires recognizing multiple interacting factors, causal chains, and synergies among them~\cite{norell1984models}. 
What ultimately makes an explanation ``good'' in medicine is its role in guiding action and responsibility~\cite{Salmon1990}. Explanations are not merely epistemic tools but also practical instruments: they justify interventions~\cite{boge2025causality}, support trust~\cite{clark2002trust}, and mediate communication between clinicians, patients, and institutions~\cite{nyrup2022explanatory,rohlfing2020explanation}. A~\textit{bona fide} medical explanation, therefore, is not simply statistically robust, but it provides the right kind of reasons in the proper context, informing treatment, justifying decisions, and sustaining trust~\cite{nyrup2022explanatory,clark2002trust}.

\section{Towards Philosophical Design Principles}
\label{sec:bridging-medical}
The disagreement over the epistemological underpinnings of XAI, together with the limited integration of philosophically grounded accounts of explanation, undermines the possibility of assessing whether explanations succeed in their epistemic, justificatory, and practical roles.
In philosophy, a medical explanation is often required to be more than descriptive: it must illuminate causal structures, justify clinical decisions, be falsifiable, and sustain the fiduciary obligations of the physician-patient relationship~\cite{russo2007interpreting,clark2002trust,qiu1989models}. Unlike explanations in other scientific fields, explanations in medicine are not only epistemic tools but also acts that distribute responsibility and secure trust in conditions of vulnerability~\cite{clark2002trust}. 
By contrast, much of XAI treats explanations instrumentally, as algorithmic artifacts primarily evaluated by metrics or user plausibility~\cite{agarwal2024faithfulness}. The consequence is an epistemic mismatch: explanations look informative but do not involve the feasibility of intervention, or they increase apparent confidence without improving decision quality~\cite{herzog2022ethical,agarwal2024faithfulness,prosperi2020causal}. While XAI can clarify decision-making processes, it may not satisfy the standards of adequacy prioritized in clinical reasoning and philosophy~\cite{holzinger2019causability}. 
Therefore, drawing on philosophical analyses of scientific explanation, we propose a set of design principles that clarify the clinical and philosophical requirements of explanations in medical AI (see Table~\ref{tab:design-principles}). 
We argue that making epistemic commitment explicit is essential if XAI explanations are to become legitimate components of clinical reasoning and accountable medical practice. 

\begin{table}[htbp]
    \centering
    \begin{tabular}{m{3cm}|m{5.8cm}|m{5.8cm}}
    \hline
    \textbf{Principle} & \textbf{Description} & \textbf{Implications for Medical XAI} \\
    \hline

    \textbf{P1: Explanations must support intervention} & Medicine is tied to intervention. Treatment aims to alter the disease.  & Post-hoc explanations must be evaluated for whether they track genuine interventionist dependencies. \\
    \hline

    \textbf{P2: Explanations are stakeholder-specific} & Medical explanations are directed toward diverse audiences, including clinicians, patients, and regulators. & XAI requires adaptive frameworks that can present causal, statistical, or narrative content tailored to the audience. \\
    \hline

    \textbf{P3: Trust must be explicitly assessed} & Empirical assessment of whether explanations actually enhance or distort clinical trust is essential. & XAI must incorporate protocols that discern plausibility and faithfulness. Explanations must reinforce justified trust. \\
    \hline
    \end{tabular}
    \vspace{0.8em}
    \caption{Philosophically grounded design principles for evaluating the epistemic commitment of medical AI explanations.}
    \label{tab:design-principles}
\end{table}

\paragraph{P1: Explanations must support intervention.} Prioritizing causal insight over correlative association is often necessary in intervention-oriented medical contexts, where treatments aim at altering disease progression rather than solely predicting outcomes~\cite{boge2021machine}. Philosophical theories, such as the epistemic theory~\cite{russo2007interpreting}, the concept of a ``web of causation''~\cite{qiu1989models}, and the probabilistic theory~\cite{schaffner1993discovery}, underscore that interventions target interacting networks of factors rather than isolated predictors. For XAI, this suggests that feature attribution methods or other post-hoc explanations must be evaluated not only for fidelity but also for whether they track genuine interventionist dependencies~\cite{boge2025causality,prosperi2020causal}. For intervention, methods that incorporate causal models, causal inference, or counterfactuals should be prioritized over those offering only associative transparency~\cite{prosperi2020causal}. Therefore, an explanation in medicine is most meaningful if it enables clinicians to determine actionable changes to influence patient outcomes.

\paragraph{P2: Explanations are stakeholder-specific.} Medical explanations are directed toward diverse audiences, including clinicians, patients, regulators, and institutional representatives. Each group requires a distinct form of intelligibility. Philosophical literature emphasizes that the adequacy of an explanation is determined by the needs of the recipient~\cite{rizzi1994causal,Pragmatics77VanFraassen}. For physicians, explanations must include causal and probabilistic detail sufficient to justify clinical interventions. For patients, narrative clarity and contextual plausibility are often more important~\cite{rohlfing2020explanation,paez2024axe}. Regulators require evidence that explanations support accountability, reproducibility, and auditability~\cite{herzog2022ethical}. Consequently, designing XAI for healthcare requires adaptive explanation frameworks that can present causal, statistical, or narrative content tailored to the audience, rather than ``one-size-fits-all'' explanations. This approach aligns with Rudin’s argument that interpretable-by-design models are preferable because they can be communicated effectively across different contexts without the need for post-hoc explanations~\cite{rudin2019stop}.

\paragraph{P3: Trust must be explicitly assessed.} Trust in medical contexts cannot be inferred from accuracy alone and must be explicitly assessed.
Philosophical work shows that trust in medicine is~\textit{sui generis}, rooted in fiduciary obligation, asymmetry, and vulnerability~\cite{clark2002trust,holland2017trust}. Trust cannot be reduced to reliability metrics, nor can plausible but unfaithful explanations secure it. Instead, trust must be cultivated through transparent mechanisms of accountability and validation. Empirical assessment of whether explanations actually enhance or distort clinical trust is therefore essential~\cite{herzog2022ethical}. Medical XAI should incorporate evaluation protocols that distinguish between plausibility and faithfulness, ensuring that explanations reinforce justified reliance rather than misplaced confidence~\cite{agarwal2024faithfulness}. 

\begin{figure}[!ht]
    \centering

    \begin{subfigure}{0.45\textwidth}
        \centering
        \includegraphics[width=\linewidth]{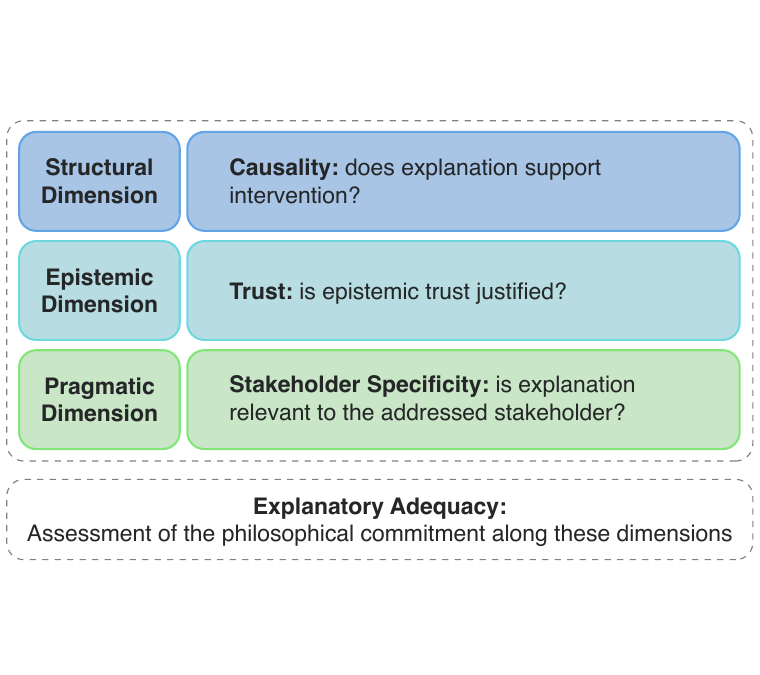}
        \caption{Evaluation dimensions for assessing XAI explanatory adequacy.}
        \label{fig:first}
    \end{subfigure}
    \hfill
    \begin{subfigure}{0.45\textwidth}
        \centering
        \includegraphics[width=\linewidth]{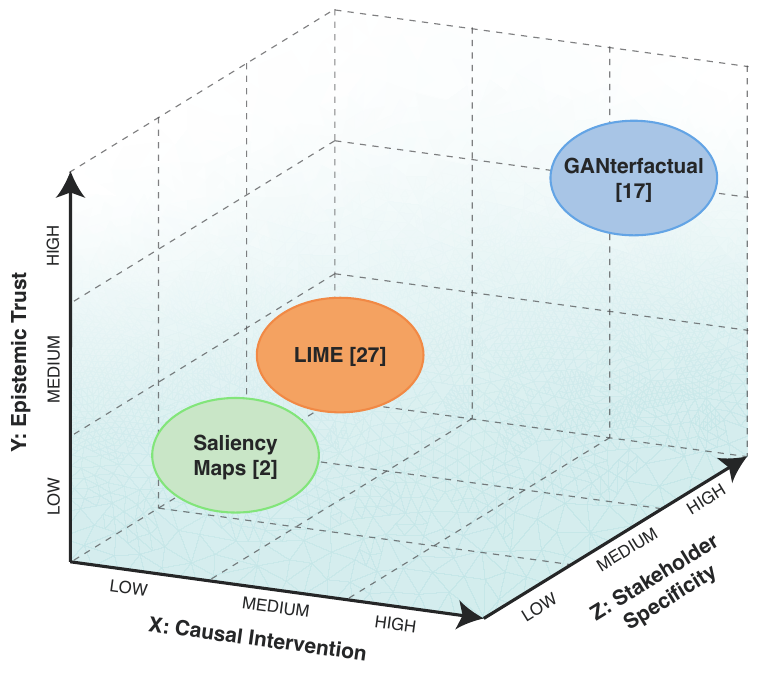}
        \caption{Positioning of analyzed XAI methods across the three dimensions of philosophical commitment.}
        \label{fig:second}
    \end{subfigure}

    \caption{Philosophical principles for assessing XAI's epistemic commitment across the dimensions of trust, causal intervention, and stakeholder specificity.}
    \label{fig:3D-principles}
\end{figure}

As Figure~\ref{fig:3D-principles} shows, these principles establish a framework for integrating~\textit{philosophically informed} medical explanations within XAI. In doing so,~\textit{they clarify the epistemic commitment of explanation, provide a basis for assessing the philosophical grounds of medical XAI methods, and help to determine how explanations are interpreted within clinical reasoning.} 
As shown in Figure~\ref{fig:second}, consider a high-accuracy breast cancer classification system, providing Grad-CAM++ heatmaps as explanations~\cite{ameen2025explainable}. The apparent effectiveness of these visualizations is often taken to entail user trust~\cite{ameen2025explainable}. However, the perceived efficacy of a saliency map does not satisfy the fiduciary notion of trust required in clinical practice: the physician, lacking access to the underlying inferential structure, is reduced to uncritical reliance on the system’s output. Crucially, saliency-based explanations can increase user confidence even when they are not faithful to the model’s actual decision process~\cite{nicolson2025human}. This becomes particularly problematic given that the heatmap conveys a spatial correlation between highlighted regions and the model’s output. For a physician, however, such a representation conveys at best correlational information, offering no causal insight into why a given region supports a diagnosis. Finally, saliency maps lack the necessary granularity and are poorly aligned with the heterogeneous epistemic needs of different stakeholders. By contrast, our framework allows us to identify the extent to which recent causal, counterfactual, or intrinsically interpretable medical AI systems satisfy these requirements.
An example is provided by a pneumonia detection system using ``GANterfactual'' explanations, in which the model generates minimally modified chest X-rays that would have resulted in a different diagnosis~\cite{mertes2022ganterfactual}. This approach provides an explicit counterfactual scenario, revealing which changes are sufficient to alter the model’s output and thereby rendering its inferential structure more intelligible in clinically relevant terms. Moreover, trust and explanation satisfaction are assessed as two separate epistemic properties through a user study explicitly involving non-expert participants, thereby addressing stakeholder-specific epistemic needs more directly than saliency-based approaches.

\section{Conclusions and Future Work}
This paper analyzed the relationship between medical explanation and XAI by reviewing the dimensions of causality, trust, and \textit{bona fide} explanations, situating current debates at the intersection of philosophical traditions and the practical demands of healthcare. 
Given the sensitive and philosophical nature of scientific explanation in medicine, XAI explanations cannot be reduced to a mere technical artifact, but should instead be understood as an epistemic practice. 
On this ground, the paper introduces the philosophical foundations of explanation as central to the development of medical XAI design principles. Accordingly, we delineate criteria for adequate medical explanations grounded in philosophical analysis.
Medicine employs multiple models of causality, depending on the context and intervention. 
The limited integration of pluralistic causal models into AI systems, largely grounded in statistical associations, limits their ability to provide the causal intervention required for clinical trust and decision-making. Indeed, trust in medicine is~\textit{sui generis}: it extends beyond simple reliability and is grounded in fiduciary responsibilities and patient vulnerability. Consequently, explanations must be evaluated for their ability to sustain the network of reliance among patients, clinicians, and technological systems. Finally,~\textit{bona fide} explanations encompass more than predictive accuracy: they need to be context-sensitive, pragmatically useful, and tailored to diverse audiences.

Indeed, a simplistic account of explanation in medicine risks overlooking the conceptual complexity of clinical reasoning. It is misleading to claim that causality is absent from medicine, thereby implying that XAI need not aspire to causal and scientific insight. Such assumptions obscure the fact that medicine has long dealt with philosophical questions about causation and explanation. Before concluding that scientific explanation is non-essential in XAI, established philosophical accounts should be revisited and integrated.
This work takes the opportunity to bridge this gap, introducing the XAI community to the rich philosophical principles that implicitly underlie many of its debates, reviving themes already discussed in the philosophy of medicine yet crucially relevant to XAI, and advancing philosophical perspectives on contemporary challenges. While this work provides a conceptual and philosophical framework, it does not systematically evaluate existing methods against the proposed explanatory criteria. Future work will address this limitation through a systematic analysis assessing how current medical XAI approaches incorporate the philosophical dimensions examined.

\bibliographystyle{plain} 
\bibliography{biblio}

\end{document}